# Applications of human activity recognition in industrial processes - Synergy of human and technology

**Friedrich Niemann**
Scientific Assistant
Chair of Materials Handling and Warehousing
TU Dortmund University

**Christopher Reining**
Chief Scientist
Chair of Materials Handling and Warehousing
TU Dortmund University

**Hülya Bas**
Scientific Assistant
Chair of Materials Handling and Warehousing
TU Dortmund University

**Sven Franke**
Scientific Assistant
Chair of Materials Handling and Warehousing
TU Dortmund University

*Human-technology collaboration relies on verbal and non-verbal communication. Machines must be able to detect and understand the movements of humans to facilitate non-verbal communication. In this article, we introduce ongoing research on human activity recognition in intralogistics, and show how it can be applied in industrial settings. We show how semantic attributes can be used to describe human activities flexibly and how context informantion increases the performance of classifiers to recognise them automatically. Beyond that, we present a concept based on a cyber-physical twin that can reduce the effort and time necessary to create a training dataset for human activity recognition. In the future, it will be possible to train a classifier solely with realistic simulation data, while maintaining or even increasing the classification performance.*

## 1. INCREASING THE EFFICIENCY OF MANUAL PROCESSES USING AUTOMATIC HUMAN ACTIVITY RECOGNITION

Increasing automation in production and logistics with simultaneous increases in the complexity of manual processes is leading to more and more interaction between humans and machines. Synergetic collaboration relies on communication, including verbal and non-verbal interactions. In material handling systems, the human factor is still a crucial variable, which is wrongly assumed to be deterministic in planning and simulation models. The time data is necessary to implement a data-driven simulation that considers the non-deterministic motion behaviour of humans. Machines must be able to detect and understand the movement of humans to facilitate non-verbal communication. One way to do this is through sensor-based human activity recognition (HAR).

HAR assign sequences of human movements recorded by sensors into a machine-readable format to predefined activities. The main advantage of this approach is its automation and scalability. In contrast to traditional methods, where movements are only recognised manually, movements can be recognised automatically. Not only simple activities but also complex applications can be detected.

HAR has already found its way into our everyday life. For example, smart watches or fitness trackers use human activity recognition to count steps, recognise types of sport, or analyse our sleep patterns. They record movements using an inertial measurement unit (IMU) and evaluate them in real-time. In addition to everyday use, HAR can be found in other domains of science and industry [1]: In health care, for example, nursing staff are automatically informed about falls of patients [2]. Another field of application is the detection of hand movements. This not only facilitates human-machine interaction [3] but also communication in the context of recognising sign language. Of the many other fields of application, the industry deserves special mention. Research is focused on production [4] and services such as intralogistics [6], [7]. While fitness trackers use HAR to distinguish between *step* and *no step*, the movements to be analysed in the industry are more complex. As movements become more complex and more detailed, automatic recognition of activities will become less accurate. The methods can be extended to include additional data streams - the so-called context [5]. A context includes information about human entities, objects and tools that do not directly involve human movements [6]. This information may relate to their condition, identity or location [7]. For example, if an employee is standing next to a shelf (object) with a scanner in hand, this data can be used to improve activity recognition [8].

In order for HAR methods to recognise movements from sensor data, they first need to be trained. Using annotated data in which the movements are already assigned to defined activities (labels), HAR methods learn to recognise different patterns in movement and context data [9], [10]. Unknown movements can then be assigned to defined activities based on these patterns. The required data is developed in the Innovationlab of the Chair of Materials Handling and Warehousing (FLW) and in cooperation with industrial partners such as MotionMiners© GmbH. The aim of recordings in such a laboratory is to examine intralogistic system at the planning phase and to record data. HAR methods are trained using this data before a system is put into operation.

*Correspondence to:* Friedrich Niemann
Chair of Materials Handling and Warehousing,
TU Dortmund University,
Joseph-von-Fraunhofer-Str. 2-4, 44227 Dortmund, Germany,
Tel. +49231 755 4548
E-mail: friedrich.niemann@tu-dortmund.de

## 2. INNOVATIONLAB FOR MOVEMENT ANALYSIS AND DATA ACQUISITION

The *Innovationlab Hybrid Services in Logistics* at FLW is a laboratory for testing innovative technologies in real-life conditions. Here, new forms of interaction between human and technology can be tested within the framework of modern logistics systems. It is the preferred environment for many research approaches due to the fact that logistics systems are physically observable in a laboratory.

First of all, even such scenarios can be reproduced in the Innovationlab without any risk of danger, which in the development phase have potentially safety-critical elements for humans. For example, the interaction of employees with transport drones, which they are supposed to perceive autonomously via various sensor interfaces, may result the risk of collision. In a real laboratory, the necessary safety precautions can be taken and additional sensor systems can be used for monitoring that would not be available in the real industrial system, such as a warehouse. Using reference sensors of the laboratory environment, solutions can be improved in a targeted manner toward practical suitability. The Innovationlab approach is fundamentally not dependent on the existence of a real existing, operational logistics system. As a result, realistic testing of technologically-driven changes is already possible during planning. For example, activity detection sensors and classifiers can be developed even for logistic systems that were not yet operational at the time of the laboratory experiments. Human activity recognition solutions that are tailored to the manual activities and processes in a newly designed or redesigned logistics system in terms of time and quantity have been available from day one. In addition, the experiments and recordings in the laboratory environment already allow conclusions to be drawn about potential improvements of the real system.

A laboratory environment provides the opportunity to assess various sensor technologies according to their applicability in specific applications of activity recognition. IMUs, video cameras and an optical motion capturing (MoCap) system are used in the Innovationlab to capture human movements. The MoCap system uses infrared cameras to capture markers worn on the body or attached to objects. Because of its high accuracy, the system serves as a reference. In this reference environment, this refers to less precise and more suitable industrial sensors. Determine the position of employees, objects example, radio frequency identification, ultra-wideband, Bluetooth low energy, WLAN or indoor global positioning systems. MoCap determines the optimal attachment of the sensors for each technology, which leads to the most meaningful context data.

## 3. ONGOING RESEARCH WORK IN THE INNOVATIONLAB

The Innovationlab is successfully used in various research projects on human activity recognition. Following is an overview that outlines their respective visions and points of contact for industrial transfer projects.

### 3.1 Attribute-based representation of classes

According to previous activity recognition research, activities such as *locomotion* and *handling* are automatically recognised by a classifier as specific activity classes. Modern Material flow systems of the vision of Industry 4.0 adapt intelligently to changing circumstances. Thus the assumption that all activity classes are known while designing a HAR solution and that their number, delimitation and definition remain the same at all times is increasingly far from reality. Consideration of new classes for each variant of the new activity complicates annotation, i.e., the labeling of the recorded data and reduces the number of examples per class. An association between a sensor pattern and a specific activity class label does not accurately project the diversity of human movements.

Semantic approaches offer a way out of the dilemma between the need for the greatest detail in the activity definition and the resulting decrease in transferability between different industrial applications on the other hand. This approach comes from image recognition [11]. Researchers assigned labels to animal images that included a semantic description of the animals, e.g., whether it was a white or black animal, and what habitat it lived in. With these attributes, a classifier is able to recognise unknown concepts or classes (in this example animals) based on their semantic description. Transferred to human activities, the idea is to represent activity classes by semantic descriptions. These descriptions are in a figurative sense letters from which words, i.e., activity classes can be flexibly formed. In [12] attributes such as *standing*, *step*, *walking*, *handling upwards*, *handling centered*, *handling downwards*, *left hand*, *right hand* as well as poses for objects such as a *cart* or *bulky* and *handy unit* are distinguished. Their combination allows the unambiguous description of arbitrarily defined classes [13]–[17].

This approach has already been successfully validated with MotionMiners©. Movement data recorded in the Innovationlab were used to train a convolutional neural network that achieved comparable activity recognition performance to a classifier trained on real systems data. In the real system, the data recording could have been used immediately for analysis. The heterogeneous activity definitions in the various systems and the laboratory data could be bridged by semantic attributes.

### 3.2 Context-aware activity recognition

As described in the previous sub chapter, words (activity classes) can be formed from semantic descriptions (attributes). But words can have different meanings, depending on how they are used in a sentence. In conclusion, the activity classes need to be put in context so that the information and its implications can be understood by humans. Through further data streams the recognised attributes can be expanded with, among other things, the identity and position of people, tools and objects and beyond that with process knowledge, state and transition logic and assigned to activity classes by a context-aware classifier.

The approach of an attribute-based representation of classes was extended by adding a random forest, which uses context information such as the position of objects (see Figure 1). The purpose of context information is to improve the performance of the convolutional neural network and to understand the captured process comprehensively. Through context information, we are able to expand the recognised activities and identify when, where and why the activity is carried out. This also allows us to deduce mistakes, such as picking the wrong item. [8]

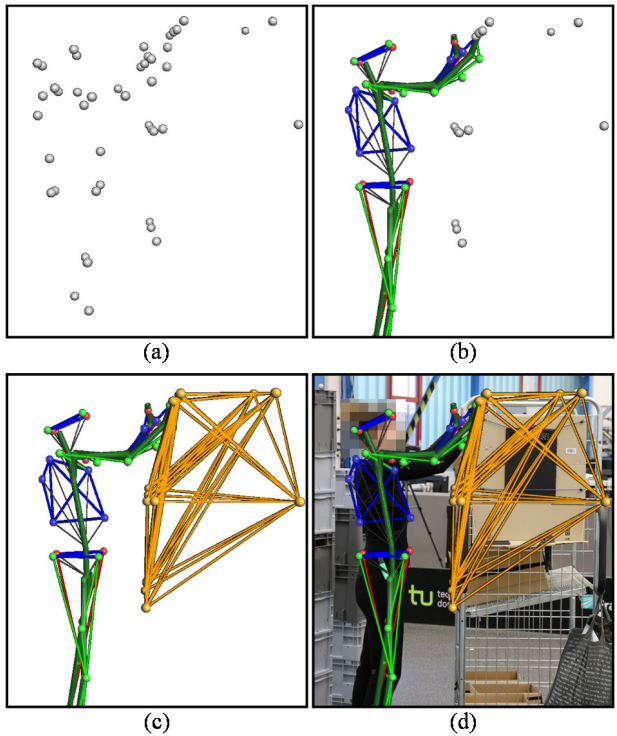

**Figure 1. Data processing of the MoCap system based on person-to-goods picking: (a) Reconstruction of a point cloud consisting of reflective markers, (b) Reconstruction of the subject based on his marker pattern, (c) Reconstruction of the picking cart, (d) Overlay of the MoCap visualisation with the video recording. [18]**

The process knowledge is used to derive the process (picking) and the sub-process (delivery confirmation). The execution time of the respective sub-processes can be collected from the data. The status and transition logic can even be used to determine previous sub-processes. Before delivery confirmation, the picker should have put at least one or more items of an order into the cart. In this step, if the result deviates from the previously determined sub-processes, the activity class can also be adjusted. The sub-process following delivery confirmation can be roughly estimated due to limitations in the transition logic. Next, two sub-processes come into question: Either the employee picks other items in his vicinity or they move forward with the picking cart, e.g. to consolidation.

The performance of activity recognition can be significantly enhanced by using context data from the Innovationlab. Furthermore, individual context features were analysed for their relevance to activity recognition [18]. By using this method, data from a warehouse can be collected in a cost- and effort-efficient manner. But context data cannot only be used for activity recognition. Process analysis and the cyber-physical twin can be conducted using the activity classes, attributes and context data.

### 3.3 Generation of human movement data - cyber-physical twin

The creation of a training dataset is associated with high effort. Data has to be recorded, annotated and then manually revised. In addition, the necessary data volume increases with an increase in complex movements, as is the case, for example, in production or intralogistics. As a result, HAR methods are mainly tested on simple everyday situations that involve the least effort in the preparation and execution of recordings. Simulations (see Figure 2) which are based on human movements, can be generated in order to reduce the necessary effort for collecting data from complex environments, such as intralogistics. Human movements are modified using sensor data from logistics and other domains resulting in synthetic and thus new movements. [19]

With the simulation environment, movements can be exported as different sensor outputs (e.g., acceleration of body regions like in an IMU or a point cloud like in a MoCap system). Depending on the data basis, the modification can even take age, gender, physical limitations or signs of fatigue. Based on these variations, person- and situation-specific simulations can be performed.

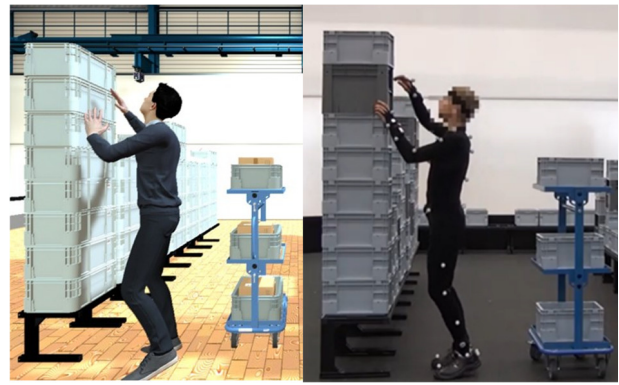

**Figure 2. Cyber-physical twin (left) of a real person-to-goods picking system (right).**

The data generation approach makes it possible to transfer planned systems to a simulation environment and to generate artificial movement data. Without the expensive and time-consuming implementation of the system in the real environment or in the laboratory, training data can be generated and HAR methods can be trained.

### 4. THE INNOVATIONLAB AS AN INTERFACE BETWEEN INDUSTRY AND RESEARCH

The conclusion of scientific research work is the empirical validation within the scope of its application. The Innovationlab with it's high-precision measuring instruments serves is an ideal reference environment. Methods, technologies, and software are analysed for their suitability for industrial application without changing ongoing operations within the company. The emphasis will be on the connection between new methods and technologies, as well as the consequences for the overall system. A variety of sensor technologies, such as the MoCap system and IMUs, were linked to simulation software and HAR methods. The research work presented in this article has been or is being validated in cooperation with various industrial partners in real warehouses. The positive experiences have confirmed the Innovationlab as an interface between industry and research.